%% file: root.tex
\documentclass[conference]{IEEEtran}
\IEEEoverridecommandlockouts

\usepackage{cite}
\usepackage{amsmath,amssymb,amsfonts}
\usepackage{algorithmic}
\usepackage{graphicx}
\usepackage{textcomp}
\usepackage{xcolor}
\usepackage{booktabs}       
\usepackage{nicematrix}
\usepackage[capitalise]{cleveref}
\usepackage[font=small]{subcaption}
\newcommand*{\ours}{RIVET}

\def\BibTeX{{\rm B\kern-.05em{\sc i\kern-.025em b}\kern-.08em
    T\kern-.1667em\lower.7ex\hbox{E}\kern-.125emX}}
\begin{document}

\title{Representation-Guided Generation and Integration of Executable Programs for Robot Manipulation
}

\author{Ruixiao Yang$^{1}$, Mingxin Yu$^{1}$, and Chuchu Fan$^{1}$
\thanks{$^{1}$Department of Aeronautics and Astronautics, Massachusetts Institute of Technology, USA, \{ruixiao, yumx35, chuchu\}@mit.edu}
}
\maketitle

\begin{abstract}
\input{tex/00-abstract}
\end{abstract}


\section{Introduction}
\label{sec: introduction}

\input{tex/01-intro}

\section{Related Work}
\label{sec: related work}
\input{tex/02-related}

\section{Methodology}
\label{sec: methodology}
\input{tex/04-method}

\section{Experiments}
\label{sec: experiments}
\input{tex/06-exp}

\section{Conclusion and Discussion}
\label{sec: conclusion}
\input{tex/08-conclusion}

\bibliography{ref}
\bibliographystyle{abbrv}

\end{document}

%% file: tex/00-abstract.tex
Building a robotic manipulation system requires connecting perception, planning, and control through carefully designed representations and interfaces. VLM code generation offers a way to automate this construction, but independently generated components may operate on incompatible geometric and task-level information.
We present \textbf{R}epresentation-guided \textbf{I}ntegration of \textbf{V}LM-generated \textbf{E}xecutable \textbf{T}ask programs (\ours{}), a framework for generating complete manipulation systems around a shared object-centric representation. The representation combines per-object 6D poses, which preserve the metric information required for action grounding, with a relation graph that exposes the task-level structure required for planning. Guided by this representation, a VLM generates cooperating perception, rendering, relation-inference, and planning programs, each combining task-specific computation with available packages where useful.
The resulting programs are authored once for a manipulation domain and reused on unseen start and goal configurations without code regeneration.
We evaluate \ours{} on cube stacking, tangram rearrangement, and three-dimensional assembly in simulation and on a physical robot, where we achieve $83\%$ overall success rate in the real world by reusing offline-generated systems. Our results demonstrate that representation-guided program generation can
adapt a common manipulation framework to tasks with different geometric, relational, and sequential requirements.

%% file: tex/01-intro.tex
\begin{figure*}[t]
    \centering
    \includegraphics[width=0.93\linewidth]{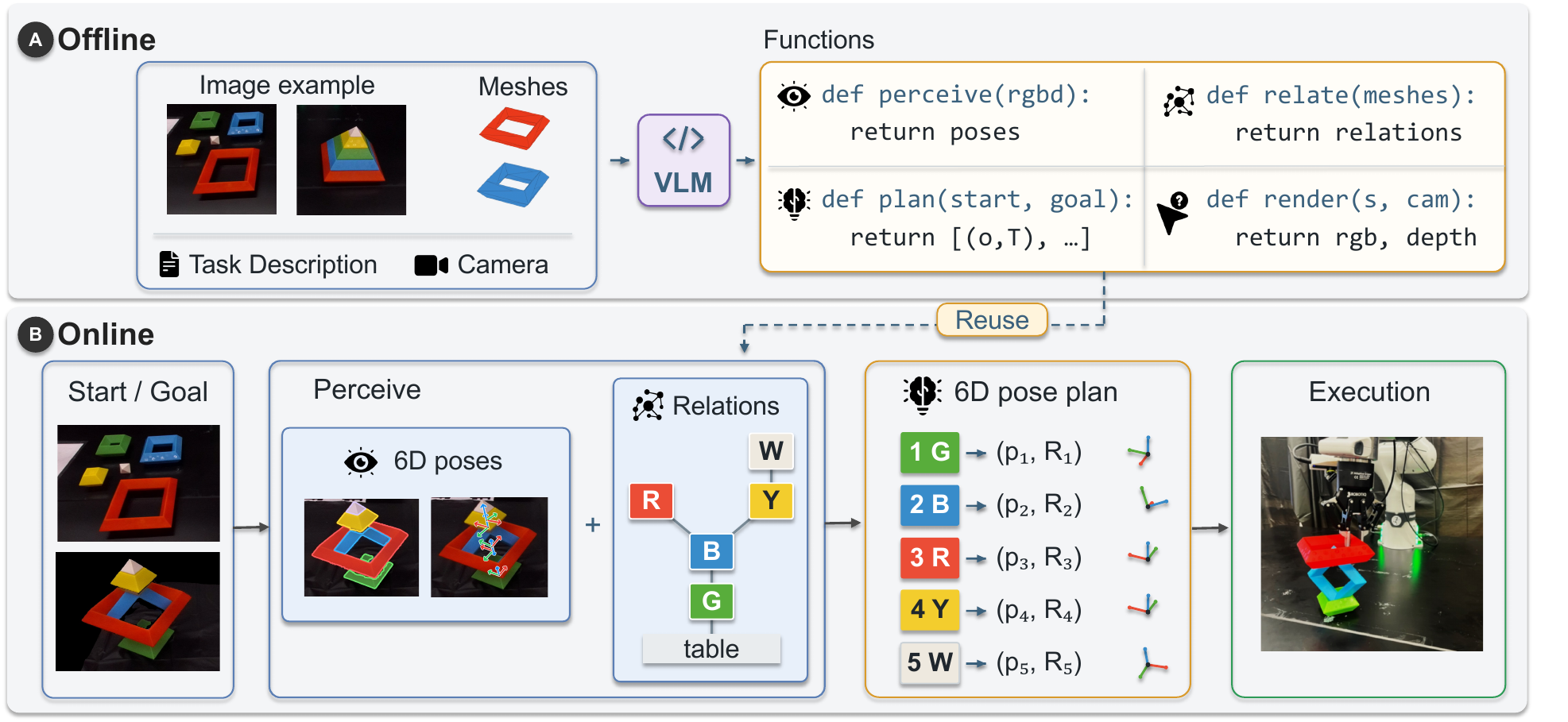}
    \caption{\textbf{Pipeline of \ours{}.}
    The system works under the guidance of an object-centric representation combining metric information from per-object 6D poses and structural information from a relation graph, which specifies the dependency of object arrangement.
    At \textit{Offline stage} (A), a VLM uses task examples, object meshes, and available tools to write reusable functions: \texttt{perceive} and \texttt{relate} transform image observations into scene representations, \texttt{plan} generates position-grounded plans to accomplish the task, and \texttt{render} maps the representation back to an image to facilitate the generation of the other functions.
    At \textit{Online stage} (B), the offline-generated \texttt{perceive}, \texttt{relate}, and \texttt{plan} functions process a new start-goal instance to generate a grounded, executable plan.
    The authored functions are reused across instances without instance-specific rewriting.}
    \label{fig:pipeline}
\end{figure*}

Vision-language models (VLMs) are becoming increasingly capable and useful for robotic manipulation. Their ability to reason about task requirements and visual observations makes them a strong fit to bridge high-level task specifications and robot behavior~\cite{driess2023palm, hu2023look, brohan2023rt}. 
However, using VLMs directly as planners makes it hard to know when and why VLMs succeed or fail. Errors in scene understanding and action reasoning can remain implicit in model responses, making failures difficult to diagnose and correct without being explicitly grounded in robotic knowledge~\cite{huang2022inner, ekpo2024verigraph}. 
Moreover, using VLM-based planner directly inside the loop can introduce substantial computational cost and high inference latency during execution.

On the other hand, traditional robotic manipulation provides useful tools within a sense-plan-act structure~\cite{siciliano2008springer}, from decades of effort in computer vision and planning. In this structure, each component solves a well-defined problem, making its behavior easier to verify and diagnose. Once suitable tools are connected through well-specified representations, the resulting system can work reliably across different initial and goal configurations at a lower cost than VLM-based planners. However, selecting these tools and designing their connections requires substantial human effort and domain expertise for each new task.

The code-generation capability of VLMs suggests a way to automate the construction of manipulation systems. Prior work has used language models to compose perception and control APIs into robot programs~\cite{liang2023code, huang2023instruct2act}, as well as to generate plans or planning models over symbolic representations~\cite{silver2024generalized, hao2026simulation}. However, generating individual program components does not by itself determine how a complete system should represent, exchange scene information and be constructed. Perception produces geometric quantities such as masks and object poses, whereas planning often depends on task-level relationships and constraints. And execution requires the planned actions to be grounded back into metric poses. If these components are generated independently, they may disagree on object identities, coordinate conventions, relation semantics, or action effects. 
The central challenge is therefore to provide a structure that connects visual observations, task-level reasoning, and executions while leaving the task-specific algorithms and their composition as a system to be generated.

To address this challenge, we present \textbf{R}epresentation-guided \textbf{I}ntegration of \textbf{V}LM-generated \textbf{E}xecutable \textbf{T}ask programs (\ours{}). Our key design is an object-centric scene representation that combines per-object 6D poses with a task-relevant relation graph. 
The poses preserve the metric information needed to reconstruct scenes and ground actions, while the graph exposes the structural information needed to reason about task constraints and action order. Because both components refer to the same object instances, the representation provides a common interface across perception, relational reasoning, and planning. Guided by this representation, the VLM generates the algorithms implementing each module, combining task-specific computation with available packages where useful. The resulting modules form an executable system that is authored once for a task domain and reused across new start and goal configurations.

Concretely, \ours{} first defines a scene representation containing both object poses and an object relation graph, then generates four cooperating functions: \texttt{perceive}, \texttt{render}, \texttt{relate}, and \texttt{plan}. \texttt{perceive} combines vision tools with generated code to recover object poses. \texttt{render} reconstructs represented scenes for inspection and facilitates the generation of \texttt{relate}, which infers task-relevant relationships. \texttt{plan} uses these representations to define action preconditions and state transitions, then searches for an action sequence with concrete placement targets.
The functions are generated sequentially by the VLM using the shared scene representation, and each function is provided with the generated code of the earlier functions on which it depends. This helps them maintain consistent meanings for object identities, poses, and relationships.
The same VLM and authoring procedure can construct the complete task-specific system across different manipulation domains. Adapting \ours{} to a new domain requires only a task description, reference observations, object geometry, camera information, and access to available computational packages; it does not require a human to implement or connect new perception, relation-inference, and planning algorithms. Within each domain, the generated programs are further reused across new initial and goal configurations without another round of VLM generation. Their explicit intermediate representations and planning rules support inspection and programmatic testing. When scene interpretation and planning are fully implemented as local programs, repeated task solving requires no runtime VLM queries.


We evaluate \ours{} on three tasks: cube stacking, tangram rearrangement, and three-dimensional assembly. These tasks span different geometric relationships, and planning constraints, allowing us to examine whether the VLM can construct reusable planning systems that solve a range of task instances.
In simulation, systems generated by \ours{} achieve more reliable task completion than the baselines. By retaining both object poses and task-relevant relationships, the generated systems connect symbolic action sequences to concrete placement targets, supporting more reliable execution when all baselines fail on two sets of tasks. 
We further demonstrate the approach on a physical robot with all three tasks, where the same generated programs remain effective across different initial and goal configurations within each task domain, and even under a change of camera view. 

Our contributions are summarized as follows:
\begin{itemize}

\item We introduce \ours{}, a representation-guided architecture that connects VLM-generated perception and planning through a shared geometric--relational scene representation.

\item We develop an automated authoring procedure that enables the same VLM to instantiate complete executable systems across manipulation domains without manual task-specific programming, and to reuse the generated programs across new configurations within each domain.

\item We evaluate \ours{} on cube stacking, tangram rearrangement, and three-dimensional assembly in simulation and on a physical robot, demonstrating cross-domain system generation and within-domain program reuse.

\end{itemize}

%% file: tex/02-related.tex
\noindent\textbf{VLMs for robotic manipulation and task planning.}
Vision-language models enable robots to interpret task instructions in the context of visual observations.
ViLa~\cite{hu2023look} uses this capability to generate task plans from observations and language or image goals, incorporating visual feedback after actions.
Structured planning methods further connect visual reasoning with explicit task constraints: VeriGraph checks model-generated plans through scene-graph transformations~\cite{ekpo2024verigraph}, while PLanAR combines VLM grounding with symbolic planning and execution verification~\cite{guo2026planar}.
In these approaches, model inference remains part of interpreting and solving tasks during deployment.
Our work investigates moving this reasoning into an offline authoring stage, producing programs that process subsequent observations and compute plans without runtime VLM inference.

\noindent\textbf{Program synthesis for robotic manipulation.}
Generating executable code allows language models to express robot behavior through explicit control flow, numerical computation, and calls to robot APIs.
Code as Policies and Instruct2Act demonstrate this approach by composing supplied perception and action tools into robot programs~\cite{liang2023code, huang2023instruct2act}.
The generated code can also encode geometric objectives: VoxPoser constructs spatial value maps for motion planning~\cite{huang2023voxposer}, and ReKep represents manipulation constraints as Python functions over three-dimensional keypoints~\cite{huang2024rekep}.
These representations expose task logic for inspection and allow conventional algorithms to perform geometric computation.
Reuse is already possible within this paradigm; VoxPoser caches generated code and reevaluates it under visual feedback~\cite{huang2023voxposer}.
This motivates examining which perception and planning capabilities can be authored as programs that remain applicable across new task configurations.

\noindent\textbf{Offline construction of reusable perception and planning systems.}
The closest work separates the acquisition of task knowledge from its repeated use.
LLM-GenPlan~\cite{silver2024generalized} synthesizes reusable planning programs from supplied PDDL domains and example problems, while PDDLLM derives planning domains from a demonstration and simulation rollouts, confining language-model token expenditure to domain construction~\cite{huang2026one}.
Recent robot-programming systems also use execution feedback during authoring: RHO~\cite{elmaaroufi2026rho} optimizes policy repositories, and ASPIRE~\cite{lu2026aspire} refines programs and accumulates reusable repair knowledge; both include evaluations of fixed programs across held-out configurations.
For visually specified tasks, the deployment requirements of perception remain consequential.
VLMFP~\cite{hao2026simulation} reuses generated planning domains but invokes a VLM to translate new visual instances into planning problems.
ProgGen~\cite{tang2025programmatic} provides a closer precedent for authoring visual processing itself, synthesizing perception, dynamics, and rendering programs for video prediction.
Our focus is the coordinated construction of visual processing and goal-directed planning for physical manipulation.
Given reference task images, a task description, and object geometry, \ours{} authors a renderer, perceiver, and solver, then freezes these programs across new initial and goal configurations within the same domain.
This formulation extends offline authoring across visual grounding and planning, making their task logic available for inspection and programmatic checking while amortizing authoring cost over repeated deployment.

%% file: tex/04-method.tex
We consider image-conditioned manipulation, where each task instance comprises a start observation $I_s$, a goal observation $I_g$, a natural-language task specification $\tau$, a collection of object meshes $\mathcal{M}$, and camera calibration $C$, including intrinsic and extrinsic parameters. The objective is to produce an executable manipulation plan \(\pi=\{a_t\}_{t=1}^{T}\) that transforms the observed start configuration into the goal configuration while satisfying the requirements specified by $\tau$. Within a task domain, $\tau$, \(\mathcal{M}\), and $C$ remain fixed, whereas \(I_s\) and \(I_g\) vary across instances.

The key design of \ours{} is an object-centric scene representation that combines per-object 6D poses with a task-relevant relation graph. The poses retain the metric information required to reconstruct the scene and ground actions in the workspace, while the relation graph exposes the structural information required to reason about constraints and action order. Because both components refer to the same object instances, the representation connects geometric perception, relational reasoning, and executable planning within a common interface.

We introduce \textbf{R}epresentation-guided \textbf{I}ntegration of \textbf{V}LM-generated \textbf{E}xecutable \textbf{T}ask programs (\ours{}), a framework that uses this representation to organize the generation of a complete manipulation system. Rather than asking the VLM to determine the system decomposition and interfaces from scratch, \ours{} specifies how each module reads or modifies the shared representation. The VLM then generates the algorithms implementing these modules, including both task-specific computation and calls to available packages where useful. In particular, it generates programs that recover metric object states from observations, reconstruct those states visually, infer task-relevant relations, and plan actions over the resulting representation. Since the responsibilities and interfaces of the modules are fixed by the representation, the generated programs can be composed into a single executable pipeline and reused across new start and goal configurations.

In the following sections, we first introduce the overview of \ours{} framework then describe the generated perception and planning modules, followed by robot grounding and execution.

\subsection{Method Overview}
The task-solving system constructed by \ours{} consists of a perception module and a planning module. Given the start and goal observations, the perception module recovers their geometric and task-relevant states, while the planning module compares these states and produces a sequence of object-level actions. Because the two modules are generated separately and operate on different forms of information, they must agree on what constitutes a scene state. We therefore manually specify a shared scene representation that serves as the interface between perception and planning.

We represent a scene as
\begin{equation}
s=\left(\mathcal{O},\mathcal{R}\right), \qquad
\mathcal{O}=\{(i,k_i,T_i)\}_{i=1}^{n}, \quad T_i\in SE(3),
\end{equation}
where $i$ identifies an object instance, $k_i$ selects its canonical mesh from $\mathcal{M}$, and $T_i$ maps the mesh into the workspace frame. The relation graph $\mathcal{R}$ contains task-relevant relations of the objects, like support dependencies between the objects. 
The metric object $\mathcal O$ records retain the geometric information needed to specify concrete placement targets, while the relations $\mathcal R$ expose the discrete conditions needed to determine valid action orderings. We manually fix this representation and its interface conventions so that object identities, coordinate frames, and relation semantics remain consistent across modules. Within this interface, the generated programs determine how the required states are constructed and used for a particular task domain.

~\cref{fig:pipeline} illustrates how these modules are generated and used. During the offline stage in~\cref{fig:pipeline}(A), the VLM receives reference start and goal observations, the task specification $\tau$, object meshes $\mathcal{M}$, camera information $C$, and interfaces to the available robotic tools. Using the reference observations to assess the task and the applicability of these tools, it generates the code for perception and planning. On the perception side, the generated code selects and composes tools with the preprocessing, postprocessing, and coordinate transformations needed to recover $\mathcal{O}$ and $\mathcal{R}$. A generated renderer maps represented object states back to visual observations, providing a common visual interface for interpreting and inspecting the recovered state. On the planning side, the generated code defines actions, preconditions, and state transitions over the same scene representation. The shared interface therefore constrains independently generated algorithms to agree on the information passed between them.

During the online stage in~\cref{fig:pipeline}(B), the generated programs are fixed and applied to a new start-goal pair. The perceiver first maps each observation into the shared representation,
\begin{equation}
s_x=f_{\mathrm{perc}}(I_x,\mathcal{M},C),
\qquad x\in\{s,g\}.
\end{equation}
The planner then takes the perceived start and goal states and produces a transformation-grounded object-level plan,
\begin{equation}
\pi=f_{\mathrm{plan}}(s_s,s_g,\tau)
   =\{(i_t,T_t^{*})\}_{t=1}^{T},
\end{equation}
where each action specifies an object instance $i_t$ and its target pose $T_t^{*}$. Finally, a fixed control interface grounds these object-level actions into robot motions and executes them. New task instances change only the start and goal observations; they do not invoke another round of VLM-based program authoring.

\subsection{Perception: Constructing Planner-Ready Scene States}

To transform an image observation into the scene representation $s=(\mathcal{O},\mathcal{R})$, \ours{} authors three functions with fixed input and output interfaces: \texttt{perceive}, \texttt{relate}, and \texttt{render}. Each function is generated with the task specification, reference observations, scene-representation conventions, and documentation for the tools relevant to its role. Dividing perception into these functions makes explicit which information each function must produce and allows their outputs to be checked before the programs are reused on new instances.

The \texttt{perceive} function recovers the metric object records $\mathcal{O}$ from an RGB-D observation. Its prompt provides the reference start and goal observations, object meshes $\mathcal{M}$, camera calibration $C$, the required object-record format, and interfaces to available perception tools. The VLM is asked to determine which tools are applicable to the observed objects and to write the complete procedure for using them, including input preparation, tool invocation, output processing, and coordinate conversion. For example, the generated procedure may obtain object masks using a custom segmentation tool from OpenCV~\cite{bradski2000opencv} and use these masks, together with the corresponding meshes and depth observation, as inputs to a pose estimator such as FoundationPose~\cite{wen2024foundationpose}. Regardless of the selected tools, \texttt{perceive} must return
\begin{equation}
\mathcal{O}=\{(i,k_i,T_i)\}_{i=1}^{n}
\end{equation}
under the instance-identity and coordinate conventions defined by the shared representation.

The \texttt{render} function provides a bridge between the metric object records produced by \texttt{perceive} and the spatial reasoning implemented by \texttt{relate}. It maps an object record $\mathcal{O}$ back into an image observation. Its authoring prompt specifies the object-record structure, mesh collection $\mathcal{M}$, camera model $C$, coordinate conventions, and interface to the available rendering package. The VLM then generates code that places each canonical mesh according to its pose $T_i$ and renders the reconstructed scene from the specified camera:
\begin{equation}
\hat{I}=\texttt{render}(\mathcal{O},\mathcal{M},C).
\end{equation}
The reconstructed image allows the VLM to compare the represented scene with the reference observation, supporting inspection of the output of \texttt{perceive} and self-verification during authoring. In addition, the generated source code is included in the prompt used to author \texttt{relate}. The source code makes the coordinate transformations, mesh conventions, and camera projection explicit, while the rendered image presents the same geometric state in a visual form. Together, they help the VLM generate relation logic that is consistent with the numerical object records produced by \texttt{perceive}. Thus, \texttt{render} does not add information to the scene representation; instead, it connects its metric and relational abstractions.

The \texttt{relate} function constructs the task-relevant relation graph $\mathcal{R}$ from the metric object records $\mathcal{O}$. Its authoring prompt provides the task specification $\tau$, the relation-graph interface, the mesh collection $\mathcal{M}$, the output contract of \texttt{perceive}, and the generated source code of \texttt{render}. The renderer code makes explicit how object poses, canonical meshes, coordinate frames, and camera parameters are combined to reconstruct a scene. This gives the VLM a concrete interpretation of the metric records when generating the geometric and logical computations used by \texttt{relate}.

Based on this context, the VLM determines which relation predicates are required for the task and generates code for computing them from the perceived object states and geometry. For example, a stacking task may require support and accessibility relations, whereas an arrangement task may require relative-position or adjacency relations. When executed, the generated function returns
\begin{equation}
\mathcal{R}=(\mathcal{V},\mathcal{E})
    =\texttt{relate}(\mathcal{O},\mathcal{M}),
\end{equation}
where each node in $\mathcal{V}$ corresponds to an object instance in $\mathcal{O}$ and each typed edge in $\mathcal{E}$ represents a task-relevant relationship between instances. Because the graph uses the same instance identifiers and geometric conventions as $\mathcal{O}$, it provides the planner with task-level structure while preserving its connection to the metric poses used for action grounding.

For a new observation $I_x$, the functions are composed as
\begin{equation}
\mathcal{O}_x
    =\texttt{perceive}(I_x,\mathcal{M},C),\quad
\mathcal{R}_x
    =\texttt{relate}(\mathcal{O}_x,\mathcal{M},\tau),
\end{equation}
and return the planner-ready state
\begin{equation}
s_x=(\mathcal{O}_x,\mathcal{R}_x).
\end{equation}
The fixed function signatures ensure that object identities and coordinate conventions are preserved from tool outputs to metric records and from metric records to relational facts. Each function is checked against its required interface before being frozen and reused across new start and goal observations.

\subsection{Planning with executable transitions}
The planning module converts the perceived start and goal states $s_s$ and $s_g$ into a sequence of grounded object actions. \ours{} authors a single \texttt{plan} function for each task domain. Its prompt provides the task specification $\tau$, the scene-representation, reference start and goal states, and the source code and output contracts of the previously generated perception functions. In particular, the implementation of \texttt{relate} specifies the meanings of the nodes and edges in $\mathcal{R}$, while the output of \texttt{perceive} specifies how object identities and poses are stored in $\mathcal{O}$. The VLM is asked to generate a planning algorithm that uses both forms of information under these conventions, rather than to produce a plan only for the reference instance.

The generated \texttt{plan} function defines the task-level actions, their preconditions, their effects on the scene state, and the condition under which the goal has been reached. Let $\mathcal{A}(s,s_g)$ denote the candidate actions generated from the current and goal states, $V(s,a)=1$ means the precondition for applying action $a$ is satisfied, and $F(s,a)$ its state transition. The generated planner searches for a sequence satisfying
\begin{equation}
s_{t+1}=F(s_t,a_t),\quad
a_t\in\mathcal{A}(s_t,s_g),\quad
V(s_t,a_t)=1,
\end{equation}
and terminates when the generated goal condition $G_\tau(s_T,s_g)$ is satisfied.

The two components of the scene representation serve different roles in this process. The relation graph $\mathcal{R}$ provides the structural constraints used to determine which objects can be moved and in what order. The metric records $\mathcal{O}$ provide the object identities and target poses needed to ground the selected actions. For example, a support edge may require an obstructing object to be moved before its support, while the corresponding pose in the goal state specifies where that object should ultimately be placed. The generated planner therefore reasons over relations to determine a valid action sequence.

During online execution, the authored function is applied to the perceived start and goal states:
\begin{equation}
\pi=\texttt{plan}(s_s,s_g)
   =\big[(i_1,T_1^{*}),\ldots,(i_T,T_T^{*})\big],
\end{equation}
where $i_t$ identifies the object moved at step $t$ and $T_t^{*}\in SE(3)$ is its target pose. Because $\tau$ is fixed within a task domain, its action rules and goal conditions are incorporated when \texttt{plan} is authored and need not be interpreted again for each instance. If no sequence satisfies the generated constraints, the function returns an explicit planning failure.

The output of \texttt{plan} specifies object-level transfers rather than robot trajectories. Grasp selection and motion generation are handled by the control interface described next. This separation allows the planner to determine task-level ordering and placement targets while leaving robot-specific feasibility and execution to the downstream controller.

\subsection{Plan Grounding and Robot Execution}

The generated plan specifies a sequence of object-level actions $\pi=\{(i_t,T_t^{*})\}_{t=1}^{T}$, where $i_t$ identifies the object to move and $T_t^{*}$ specifies its target pose in the calibrated workspace frame. A fixed execution interface uses the observed and target poses under the robot base frame using the camera-to-robot calibration. Unlike the perception and planning programs, this interface is not generated by the VLM and remains unchanged across task domains.

For each action, the executor selects a grasp compatible with the object's observed and target poses, checks kinematic and geometric feasibility, and generates the corresponding approach, grasp, transfer, release, and retreat motions. These robot-side procedures determine how a planned transfer is realized without changing the object identity, target pose, or action order specified by the planner. The actions are executed sequentially without intermediate visual replanning.

%% file: tex/06-exp.tex
\begin{figure*}[t]
    \centering
    \includegraphics[width=0.93\linewidth]{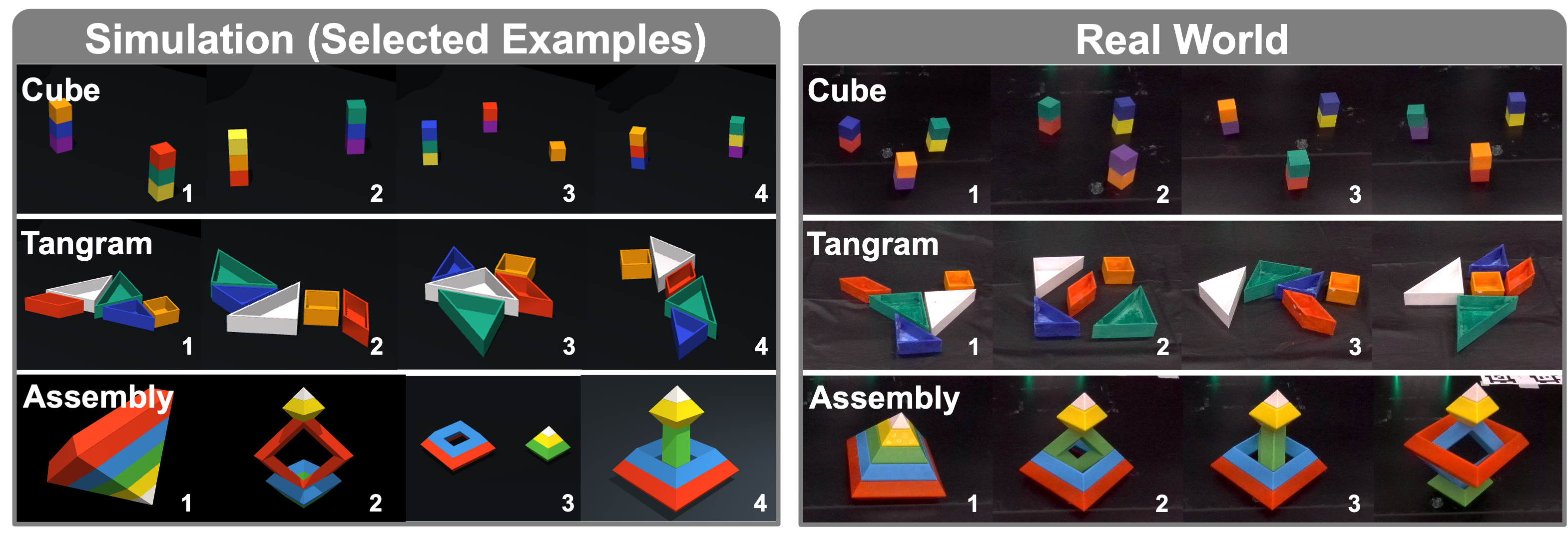}
    \caption{\textbf{Experimental task configurations used in simulation and the real world.} Rows show cube stacking, tangram arrangement, and 3D block assembly. Columns show four target configurations evaluated for each task.}
    \label{fig:exp_tasks}
\end{figure*}


We evaluate whether \ours{} can automatically construct effective and reusable manipulation systems across tasks with different geometric and structural requirements. Our experiments address two primary questions: 
(1) Can \ours{} generate systems that solve tasks with different perception and planning requirements, and how do these systems compare with online VLM planners and alternative program-generation methods?
(2) Once authored for a task domain, can a fixed generated system be reused across new initial--goal configurations without code regeneration? 
We study these questions through simulation comparisons, real-world deployment, and evaluation under camera-view changes. We further analyze where failures arise across different stages, and report the computation cost associated with program generation.

\subsection{Experimental Setup}
\label{sec:setup}

\textbf{Platform.} We use a Franka Research 3 arm with a Robotiq 2F-85 gripper to manipulate objects. A ZED M camera is mounted on the wrist, and a ZED~2i RGB-D camera is positioned in front of the arm to observe the workspace. All devices are connected to a workstation with an AMD Ryzen Threadripper PRO 5955WX 16-core and an NVIDIA GeForce RTX 4090 to accommodate grasp generation and potential neural-based vision tools in the perceiver. The VLM we used is Claude Opus 5.

\textbf{Tasks.}
We evaluate the same three task families in simulation and the real world. Each instance is specified by RGB-D observations of the current and goal configurations together with a task description. The evaluated configurations are shown in~\cref{fig:exp_tasks}. The tasks impose complementary geometric and relational demands:

\begin{enumerate}
    \item \textbf{Cube stacking:} The robot rearranges several cubes into target stacks. Because cube orientation is immaterial, the goal primarily constrains each cube's 3D position. Successful execution requires precise horizontal alignment, correct placement height, and a valid support order: cubes on top must be placed after its support. This task therefore tests metric 3D perception, placement precision, and sequential planning.

    \item \textbf{Tangram arrangement:} The robot rearranges tangram pieces to reproduce a target planar pattern. In contrast to cube stacking, this task primarily constrains each piece's planar position and orientation. The tightly packed arrangements make small translational or rotational errors visible as gaps or overlaps. Tangram thus tests piece correspondence and precise joint reasoning over planar translation and rotation.

    \item \textbf{3D block assembly:} The robot rearranges shaped components into a target three-dimensional structure. The components can severely occlude one another, making their poses and physical relationships difficult to infer from the observations. The robot must also reason about support, accessibility, and assembly order so that each component can be placed and the resulting structure remains stable. This task tests whether the framework can integrate metric perception with relational reasoning and physically grounded planning.
\end{enumerate}

\textbf{Protocol.}
In simulation, we evaluate each task on 20 goal configurations, each paired with three initial configurations, yielding 60 trials per task. We conduct 10 independent program-generation runs and evaluate each generated system on six start-goal pairs. We report success at three stages: state estimation, symbolic planning, and final execution. By evaluating multiple independently generated systems rather than a single selected program, this protocol measures the reliability of our program-generation framework. We set the runtime limit to 10 minutes for each instance.

In the real world, we evaluate four goals per task, with five trials per goal using different initial object configurations, yielding 20 trials per task. For each task, we generate one system using Goal~1, freeze it, and reuse it across all 20 trials without modification. This protocol evaluates the reliability of a fixed generated system under variations in initial configuration and physical execution, while performance on Goals~2--4 measures generalization beyond the authoring goal.

\begin{table*}[t]
\centering
\caption{
\textbf{Planning result in simulation.} Each column is tested with 60 trials. We report the success rate based on three crucial stages. \textbf{State AC} is the ratio of accurate acquisition of the state information required by the downstream planner (VeriGraph evaluates the relation graph only, while \ours{} uses both pose and relation). \textbf{Plan SR} is the success rate of generating a correct plan symbolically. \textbf{Final SR} is the success rate of controlling the robot to manipulate from the start scenario to the goal scenario in the MuJoCo simulator. ``N/A'' denotes that the metric is not applicable to the evaluated method, where ViLa, CaP, Instruct2Act omit the state and generate the plan directly.}
\label{tab:planning_execution_results}
\input{tables/simulation}
\end{table*}

\textbf{Baselines.}
To compare our methods, we compare our method against:
\paragraph{Models that plan during execution}
We compare with ViLa~\cite{hu2023look} and VeriGraph~\cite{ekpo2024verigraph}.
VeriGraph uses a VLM to construct scene graphs from an initial image and a goal specified by an image or language instruction. An LLM receives these graphs, an action set, and task constraints to predict an action sequence. The proposed actions are applied as graph edits and checked for constraint violations. The updated graph and any detected violations are returned to the planner to refine subsequent predictions.
ViLa directly prompts a VLM with the current scene image and task instruction, optionally including a reference goal image, to predict a sequence of actions. The robot executes the first action using an available primitive skill. The updated observation and action history are then provided to the VLM to generate the next plan. This process repeats until the model indicates task completion.

\paragraph{Models that write reusable programs}
We compare with Code as Policies (CaP)~\cite{liang2023code} and Instruct2Act~\cite{huang2023instruct2act}.
CaP translates natural-language instructions into executable Python programs that queries available perception APIs, performs geometric and spatial reasoning, and invokes predefined robot skills to accomplish the task.
Instruct2Act uses an LLM to generate executable programs that compose predefined perception and manipulation APIs. Following the original method, we provide image-level perception tools SAM~\cite{kirillov2023segment} for object segmentation and CLIP~\cite{radford2021learning} for identification, and the generated program reasons over their outputs to construct pick-and-place actions. We preserve Instruct2Act's original image-level reasoning interface, while adapting its pick-and-place primitive to our robot environment.

\subsection{Main comparison in simulation}
\label{sec:sim_main}

~\cref{tab:planning_execution_results} reports performance at three stages. \ours{} achieves final success rates of 100\% on cube stacking and 90\% on tangram arrangement. These tasks aims for complementary capabilities: cube stacking for accurate placement and support-aware action ordering, tangram arrangement for precise planar positioning and orientation. Three-dimensional assembly further challenges the system to recover partially occluded geometry and reason about support and assembly order.

The intermediate metrics help distinguish successful task reasoning from successful physical completion. An incorrect scene representation can lead a planner to solve the wrong problem, while a symbolically valid plan can still fail if its actions are not grounded in appropriate placement poses. The shared representation in \ours{} makes both geometric targets and task-relevant relationships available to the generated planning code. Evaluating the complete pipeline therefore tests whether these components work together to produce executable behavior.

\subsection{Comparative Analysis and Failure Modes}
\label{sec:failure_analysis}

\paragraph{Baseline failures.}
The baselines expose complementary limitations at the interface between observation and planning. ViLa reasons about scene structure implicitly and frequently misses action dependencies, such as occupied destinations, supports that must first be cleared, and the need for temporary placements. VeriGraph makes these dependencies explicit, but errors in its inferred relations produce incorrectly specified planning problems. Moreover, even a correct relation graph does not preserve metric object poses for planning. For example, globally translating all objects leaves their pairwise relations unchanged but requires every object to be moved. Code-generation baselines leave the internal scene representation to the generated program. Both CaP and Instruct2Act tend to encode task relations using simplified heuristics. For execution, Instruct2Act generates pixel-level actions without estimating object poses. Because image coordinates do not uniquely specify depth or a pose in the robot frame, these actions cannot directly provide consistent 3D placement targets.

\paragraph{What \ours{} addresses.}
These failures expose the need to retain both geometric and structural scene information. \ours{} represents each object by a 6D pose for metric action grounding and includes a relation graph for reasoning about dependencies and action order. This allows the planner to use relations to determine which actions are valid while obtaining concrete placement targets from their poses. The representation-guided function interfaces further
ensure that perception and planning use consistent object identities,
coordinate conventions, and relation semantics. The intermediate outputs also allow us to attribute an unsuccessful trial to state estimation, relational reasoning, planning, or execution.

\paragraph{Remaining failures of \ours{}.}
\ours{} does not eliminate errors within the generated algorithms or the underlying robotic components. Its main perception failures occur under severe occlusion, where the generated perceiver may select an incorrect pose hypothesis for a partially visible tangram or assembly component. Some generated planners also exceed the planning budget because their search procedures explore the state space inefficiently. Physical execution can additionally fail because of grasp or motion feasibility. Thus, \ours{}
addresses the consistency and integration of perception and planning, but its
performance remains bounded by the quality of the selected perception
methods, generated search algorithms, and robot controller.

\subsection{Applicability in Real World}
We report the performance of our framework over 60 real-world trials in~\cref{tab:exp-main}. \ours{} succeeds in 18/20 trials for both cube stacking and tangram arrangement and in 14/20 trials for 3D assembly, yielding an overall success rate of 50/60 (83.3\%). The lower assembly success rate reflects the added difficulty of jointly satisfying 3D pose, support, and motion-feasibility constraints during physical execution. Importantly, the task-domain programs are generated using Goal~1, then frozen and reused across all four goals without code regeneration. Their performance on Goals~2--4 demonstrates generalization beyond the authoring configuration.

\begin{table}[t]
    \centering
    \small
    \caption{\textbf{Real-world execution results.} Each entry reports the number of success out of five attempts. For each task domain, the VLM-authored programs are generated using Goal~1, then frozen and reused across all four goals.}
    \label{tab:exp-main}
    \input{tables/real}
\end{table}

\subsection{Robustness to Camera Viewpoint}
\label{sec:exp_ablations}

\begin{table}[t]
    \centering
    \caption{\textbf{Robustness under camera pose change.}}
    \label{tab:component}
    \small
    \input{tables/ablation}
\end{table}

We further evaluate sensitivity to camera viewpoint in~\cref{tab:component}. 
We directly test our generated system with two different camera views. Since the authorized system takes the camera as a parameter, adapting a new camera view is directly applicable without any modification in the code.

The results suggest that the generated systems can operate under the tested viewpoint changes, with performance depending on the visibility of task-relevant objects and relationships. Views that introduce substantial occlusion make pose recovery more difficult and can consequently affect relation inference and planning.
This behavior is consistent with the role of the object-centric representation: the planner operates on recovered poses and relationships, while the perceiver handles changes in image appearance. Viewpoint changes can therefore be accommodated when the scene representation remains accurate. The remaining sensitivity to occlusion identifies perception as an important limitation of this transfer.

%% file: tables/simulation.tex
\setlength{\tabcolsep}{10pt}
\begin{NiceTabular}{lccccccccc}
\toprule
& \multicolumn{3}{c}{\textbf{Cubes}}
& \multicolumn{3}{c}{\textbf{Tangram}}
& \multicolumn{3}{c}{\textbf{Assembly}} \\
\cmidrule(lr){2-4}
\cmidrule(lr){5-7}
\cmidrule(lr){8-10}
\textbf{Method}
& \textbf{State AC}
& \textbf{Plan SR}
& \textbf{Final SR}
& \textbf{State AC}
& \textbf{Plan SR}
& \textbf{Final SR}
& \textbf{State AC}
& \textbf{Plan SR}
& \textbf{Final SR} \\
\midrule
ViLa
& N/A & 43.3\% & N/A 
& N/A & 30.0\% & N/A 
& N/A & 8.3\% & N/A \\

VeriGraph
& \textbf{100\%} & 86.7\% & 63.3\%
& 80.0\% & 33.3\% & 0.0\%
& 10.0\% & 10.0\% & 0.0\% \\

CaP
& N/A & 86.7\% & N/A
& N/A & 60.0\% & N/A
& N/A & 83.3\% & N/A \\

Instruct2Act
& N/A & 60.0\% & 46.7\%
& N/A & \textbf{96.7\%} & 0.0\%
& N/A & 0.0\% & 0.0\%\\

\midrule
\textbf{\ours{}(Ours)}
& \textbf{100\%} & \textbf{100\%} & \textbf{100\%}
& \textbf{90.0\%} & 90.0\% & \textbf{90.0\%}
& \textbf{93.3\%} & \textbf{86.7\%} & \textbf{86.7\%}\\
\bottomrule
\end{NiceTabular}

%% file: tables/real.tex
\setlength{\tabcolsep}{7pt}
\begin{tabular}{lccccc}
    \toprule
    Task & Goal 1 & Goal 2 & Goal 3 & Goal 4 & Total\\
    \midrule
    Cube & 5/5 & 4/5 & 4/5 & 5/5 & 18/20\\
    Tangram & 5/5 & 5/5 & 5/5 & 3/5 & 18/20 \\
    Assembly & 3/5 & 4/5 & 4/5 & 3/5 & 14/20 \\
    \bottomrule\\
\end{tabular}

%% file: tables/ablation.tex
\begin{tabular}{lccc}
    \toprule
    Tasks & Original & Viewpoint 1 & Viewpoint 2 \\
    \midrule
    Cube Stacking & 100\% & 93.3\% & 100\% \\
    Tangram & 90\% & 100\% & 86.7\% \\
    3D block assembly & 86.7\% & 80\% & 76.7\% \\
    \bottomrule
\end{tabular}

%% file: tex/08-conclusion.tex
We presented \ours{}, an abstraction-guided VLM tool orchestration framework that generates task-specific perception, rendering, and planning programs around shared geometric and relational representations. The generated programs combine specialized robotic tools with task-specific processing and can be reused across new initial and goal configurations without repeated VLM-based authoring. We evaluate this framework on cube stacking, tangram rearrangement, and three-dimensional assembly in simulation and on a physical robot. Future work will incorporate execution feedback and failure-recovery replanning while extending the framework to more complex, contact-rich manipulation tasks.

%% file: ref.bib
@inproceedings{wen2024foundationpose,
  title={Foundationpose: Unified 6d pose estimation and tracking of novel objects},
  author={Wen, Bowen and Yang, Wei and Kautz, Jan and Birchfield, Stan},
  booktitle={2024 IEEE/CVF Conference on Computer Vision and Pattern Recognition (CVPR)},
  pages={17868--17879},
  year={2024},
  organization={IEEE}
}

@article{guo2026planar,
  title={PLanAR: Planning-Language-Grounded Agentic Reasoning for Robot Manipulation},
  author={Guo, Pengyuan and Mai, Zhonghao and Xu, Zhengtong and Zhang, Kaidi and Luu, Quan Khanh and Zhang, Heng and Miao, Zichen and Ajoudani, Arash and Kingston, Zachary and Qiu, Qiang and She, Yu},
  journal={arXiv preprint arXiv:2602.01662},
  year={2026}
}

@article{hu2023look,
  title={Look before you leap: Unveiling the power of gpt-4v in robotic vision-language planning},
  author={Hu, Yingdong and Lin, Fanqi and Zhang, Tong and Yi, Li and Gao, Yang},
  journal={arXiv preprint arXiv:2311.17842},
  year={2023}
}

@article{ekpo2024verigraph,
  title={Verigraph: Scene graphs for execution verifiable robot planning},
  author={Ekpo, Daniel and Levy, Mara and Suri, Saksham and Huynh, Chuong and Swaminathan, Archana and Shrivastava, Abhinav},
  journal={arXiv preprint arXiv:2411.10446},
  year={2024}
}

@inproceedings{liang2023code,
  title={Code as policies: Language model programs for embodied control},
  author={Liang, Jacky and Huang, Wenlong and Xia, Fei and Xu, Peng and Hausman, Karol and Ichter, Brian and Florence, Pete and Zeng, Andy},
  booktitle={2023 IEEE International conference on robotics and automation (ICRA)},
  pages={9493--9500},
  year={2023},
  organization={IEEE}
}

@article{huang2023voxposer,
  title={Voxposer: Composable 3d value maps for robotic manipulation with language models},
  author={Huang, Wenlong and Wang, Chen and Zhang, Ruohan and Li, Yunzhu and Wu, Jiajun and Fei-Fei, Li},
  journal={arXiv preprint arXiv:2307.05973},
  year={2023}
}

@article{huang2023instruct2act,
  title={Instruct2act: Mapping multi-modality instructions to robotic actions with large language model},
  author={Huang, Siyuan and Jiang, Zhengkai and Dong, Hao and Qiao, Yu and Gao, Peng and Li, Hongsheng},
  journal={arXiv preprint arXiv:2305.11176},
  year={2023}
}

@article{huang2024rekep,
  title={Rekep: Spatio-temporal reasoning of relational keypoint constraints for robotic manipulation},
  author={Huang, Wenlong and Wang, Chen and Li, Yunzhu and Zhang, Ruohan and Fei-Fei, Li},
  journal={arXiv preprint arXiv:2409.01652},
  year={2024}
}

@inproceedings{silver2024generalized,
  title={Generalized planning in pddl domains with pretrained large language models},
  author={Silver, Tom and Dan, Soham and Srinivas, Kavitha and Tenenbaum, Joshua B and Kaelbling, Leslie and Katz, Michael},
  booktitle={Proceedings of the AAAI conference on artificial intelligence},
  volume={38},
  number={18},
  pages={20256--20264},
  year={2024}
}

@inproceedings{huang2026one,
  title={One Demo Is All It Takes: Planning Domain Derivation with LLMs from A Single Demonstration},
  author={Huang, Jinbang and Xiao, Yixin and Zhang, Zhanguang and Coates, Mark and Hao, Jianye and Zhang, Yingxue},
  booktitle={International Conference on Learning Representations},
  volume={2026},
  pages={2489--2530},
  year={2026}
}

@article{elmaaroufi2026rho,
  title={RHO: Your Coding Agent is Secretly a Roboticist},
  author={Elmaaroufi, Karim and Svegliato, Justin and Kalade, Sarunas and Schelle, Graham and Seshia, Sanjit A and Zaharia, Matei},
  journal={arXiv preprint arXiv:2606.16458},
  year={2026}
}

@article{lu2026aspire,
  title={ASPIRE: Agentic/Skills Discovery for Robotics},
  author={Lu, Runyu and Wu, Yubo and Kou, Ethan and Fu, Letian and Xiao, Wenli and Mandlekar, Ajay and Xu, Yinzhen and Shi, Guanya and Goldberg, Ken and Chen, Ang and others},
  journal={arXiv preprint arXiv:2607.00272},
  year={2026}
}

@inproceedings{hao2026simulation,
  title={Simulation to rules: A Dual-VLM framework for formal visual planning},
  author={Hao, Yilun and Chen, Yongchao and Fan, Chuchu and Zhang, Yang},
  booktitle={International Conference on Learning Representations},
  volume={2026},
  pages={50993--51032},
  year={2026}
}

@article{tang2025programmatic,
  title={Programmatic Video Prediction Using Large Language Models},
  author={Tang, Hao and Ellis, Kevin and Lohit, Suhas and Jones, Michael J and Chatterjee, Moitreya},
  journal={arXiv preprint arXiv:2505.14948},
  year={2025}
}

@article{driess2023palm,
  title={Palm-e: An embodied multimodal language model},
  author={Driess, Danny and Xia, Fei and Sajjadi, Mehdi SM and Lynch, Corey and Chowdhery, Aakanksha and Ichter, Brian and Wahid, Ayzaan and Tompson, Jonathan and Vuong, Quan and Yu, Tianhe and others},
  journal={arXiv preprint arXiv:2303.03378},
  year={2023}
}

@book{siciliano2008springer,
  title={Springer handbook of robotics},
  author={Siciliano, Bruno and Khatib, Oussama and Kr{\"o}ger, Torsten},
  volume={200},
  year={2008},
  publisher={Springer}
}

@article{huang2022inner,
  title={Inner monologue: Embodied reasoning through planning with language models},
  author={Huang, Wenlong and Xia, Fei and Xiao, Ted and Chan, Harris and Liang, Jacky and Florence, Pete and Zeng, Andy and Tompson, Jonathan and Mordatch, Igor and Chebotar, Yevgen and others},
  journal={arXiv preprint arXiv:2207.05608},
  year={2022}
}

@article{brohan2023rt,
  title={Rt-2: Vision-language-action models transfer web knowledge to robotic control},
  author={Brohan, Anthony and Brown, Noah and Carbajal, Justice and Chebotar, Yevgen and Chen, Xi and Choromanski, Krzysztof and Ding, Tianli and Driess, Danny and Dubey, Avinava and Finn, Chelsea and others},
  journal={arXiv preprint arXiv:2307.15818},
  year={2023}
}

@inproceedings{radford2021learning,
  title={Learning transferable visual models from natural language supervision},
  author={Radford, Alec and Kim, Jong Wook and Hallacy, Chris and Ramesh, Aditya and Goh, Gabriel and Agarwal, Sandhini and Sastry, Girish and Askell, Amanda and Mishkin, Pamela and Clark, Jack and others},
  booktitle={International conference on machine learning},
  pages={8748--8763},
  year={2021},
  organization={PmLR}
}

@INPROCEEDINGS{kirillov2023segment,
  author={Kirillov, Alexander and Mintun, Eric and Ravi, Nikhila and Mao, Hanzi and Rolland, Chloe and Gustafson, Laura and Xiao, Tete and Whitehead, Spencer and Berg, Alexander C. and Lo, Wan-Yen and Dollár, Piotr and Girshick, Ross},
  booktitle={2023 IEEE/CVF International Conference on Computer Vision (ICCV)}, 
  title={Segment Anything}, 
  year={2023},
  volume={},
  number={},
  pages={3992-4003},
  doi={10.1109/ICCV51070.2023.00371}}

@article{bradski2000opencv,
  title={OpenCV},
  author={Bradski, Gary and Kaehler, Adrian and others},
  journal={Dr. Dobb’s journal of software tools},
  volume={3},
  number={2},
  pages={1--81},
  year={2000}
}
